\documentclass[a4paper, twocolumn]{article}
\usepackage{graphicx}
\usepackage{latexsym}
\usepackage{cite}
\usepackage{amsmath,amssymb,amsfonts}
\usepackage{textcomp}
\usepackage{xcolor}
\usepackage{booktabs}
\usepackage{tabularx}
\usepackage{subfigure}
\usepackage{hyperref}
\usepackage{algorithm}
\usepackage{algpseudocode}

\usepackage{refcount}
\usepackage{soul}
\usepackage[misc]{ifsym}

\title{\emph{AttentionMix}: Data augmentation method that relies on BERT attention mechanism}

\date{}
\begin{document}

\author{Dominik Lewy, 
Jacek Ma{\'n}dziuk  \\
Faculty of Mathematics and Information Science, \\
Warsaw University of Technology, \\
\tt{dominik.lewy@gmail.com, jacek.mandziuk@pw.edu.pl}\\}



\twocolumn[
  \begin{@twocolumnfalse}
    \maketitle
    \begin{abstract}
      The \textit{Mixup} method has proven to be a powerful data augmentation technique in Computer Vision, with many successors that perform image mixing in a guided manner. One of the interesting research directions is transferring the underlying \textit{Mixup} idea to other domains, e.g. Natural Language Processing (NLP). Even though there already exist several methods that apply \textit{Mixup} to textual data, there is still room for new, improved approaches. In this work, we introduce \textit{AttentionMix}, a novel mixing method that relies on attention-based information. While the paper focuses on the BERT attention mechanism, the proposed approach can be applied to generally any attention-based model. \textit{AttentionMix} is evaluated on 3 standard sentiment classification datasets and in all three cases outperforms two benchmark approaches that utilize \textit{Mixup} mechanism, as well as the vanilla BERT method. The results confirm that the attention-based information can be effectively used for data augmentation in the NLP domain.
    \end{abstract}
  \end{@twocolumnfalse}
]


\section{Introduction}
In recent years the introduction of transformer architectures~\cite{attention} dominated many academic tasks (e.g.~\cite{GLUE}) and commercial applications in the Natural Language Processing (NLP) area, with Bidirectional Encoder Representations from Transformers (BERT)~\cite{BERT} architecture having a special position in this genre.

In parallel, an interesting group of data augmentation methods that rely on mixing images was introduced and studied in-depth in the Computer Vision (CV) domain. The canonical method in this group is \textit{Mixup}~\cite{mixup}. The method mixes images and their corresponding one-hot-encoded labels linearly during the training process to create synthetic observations. This method has many successors that adjusted it to specific settings \cite{mix_style,snap_mix}, addressed particular vulnerabilities~\cite{manifold_mixup} or made the mixing process more effective by means of a certain guidance~\cite{saliency_mix,attentive_cutmix,snap_mix}. 

The concept of mixing training samples is also applicable to the NLP domain for sentence classification~\cite{mixup_for_text,mixup_for_text_2}. However, in this area it is not yet well studied, with only several methods that try to apply the vanilla \textit{Mixup} approach.

In this work, we propose an extension of the \textit{Mixup} method for sentence classification that uses the information on attention~\cite{attention} coming from BERT model to improve the sentence mixing procedure. 
We show that the proposed approach outperforms the base BERT method and two methods that apply vanilla \textit{Mixup}. Additionally, we perform an ablation study to explain why certain attention layers produce more meaningful information, from the point of view of the training process, that results in higher accuracy for a given task.
%

\subsection{Motivation}
Data augmentation techniques differ significantly between modalities, e.g. CV and NLP. In CV, there is a multitude of relatively simple data augmentation mechanisms like rotation, cropping, flipping, scaling, etc. The vast majority of them are modality-specific and are not applicable to text. In NLP, the number of relatively simple augmentation mechanisms is limited to just a few, such as synonym replacement or random insertion/deletion/swap~\cite{text_simple_aug}.

The main motivation of this work is to extend the repertoire of relatively simple augmentation methods in the NLP domain by enabling the use of mixing augmentation methods that proven very effective in CV. To this end, we enhance the \textit{Mixup} method by adding text-specific attention-based mechanism.

\subsection{Contribution}
The main contribution of this work is a proposal of a new augmentation method for sentence classification (\textit{AttentionMix}) based on the \textit{Mixup} method, which utilizes attention weights to guide the mixing process. Even though the paper focuses on attention utilization in BERT~\cite{BERT}, the same approach can be applied to any other architecture that implements attention mechanism~\cite{attention}. \emph{AttentionMix} is empirically tested on $3$ sentence classification datasets. Additionally, the information stored in the best performing attention layers is analyzed to shed light on the reasons of their strong performance.

While in the literature there have already been proposed several guided augmentation approaches transferred from the CV domain to the NLP domain (summarized in Section~\ref{sec:Mixup-like}), this paper is the first to propose a guided text augmentation method rooted in the NLP-specific mechanism, i.e. attention.

\section{Related work}

\subsection{\textit{Mixup} like data augmentation}\label{sec:Mixup-like}

A canonical method that started the research on image mixing as a form of data augmentation is \textit{Mixup}~\cite{mixup}, that linearly interpolates two images and its corresponding labels. There are other methods that perform mixing in a guided manner, i.e. by means of identifying the most relevant parts of the image, e.g.~\cite{attentive_cutmix} which uses information coming from a CNN classifier, \cite{saliency_mix} that applies statistical approach, or~\cite{snap_mix} which, in the mixing process, utilizes the neural network gradients. A review of mixing based data augmentation techniques for image classification is presented in the recent survey paper~\cite{naszeSurvey}.

The above idea, that originated in the CV domain, proven also useful in text classification. \cite{mixup_for_text} utilizes the \textit{Mixup} idea to perform augmentation of word embeddings and sentence embedding in the training process of CNN and LSTM networks. Subsequent works~\cite{mixup_for_text_2,mixup_for_text_3} extend this research by considering BERT architecture and experimenting with \textit{Manifold Mixup}~\cite{manifold_mixup} -- a variation of \textit{Mixup}, which applies mixing to hidden layers of the network. Another example is~\cite{ssmix} which utilizes gradient based saliency information to mix the original sentence on a word level. \textit{DropMix}~\cite{dropmix}, also utilizes gradient based saliency information, but additionally combines mixing with dropout mechanism to obtain a mixed sample.

All the above methods verify whether the mechanisms that have already been successful in CV can be effectively applied to NLP, albeit with certain domain-specific adjustments (e.g. changing a sentence into an embedding that can be mixed~\cite{mixup_for_text,mixup_for_text_2} or summing gradient based saliency information at the word level~\cite{ssmix}). 

Contrary to the above approaches that rely on improvements rooted in the CV domain, our method is the first to use the guidance stemming from the text-specific mechanism, i.e. attention.

\subsection{Methods utilizing attention} 
An interesting example of this type of method is~\cite{attention_annotation} which utilizes attention from BERT as a means to determine important words in pregnancy-related questions for annotation purposes. The approach does not differentiate attention layers/heads, and for a given word simply aggregates attention from all heads from all layers, and deems words with higher attention sum as more important. A very similar approach is proposed in~\cite{attention_hate_speach} to identify keywords indicative of hate speech. An interesting finding from this paper is that for different datasets (Boomer-hate vs. Asian-hate) the attention mechanism behaves differently (in the former, it associates hate words with a target group, and in the latter, it does not), which suggest that the applicability of attention may differ between datasets.

There are also quite many papers that analyse the properties of the attention weights. For instance, \cite{annotation_analysis} tries to understand attentions in the context of their relative position or density, as well as from the perspective of whether particular attention layers/heads learned any language structures (e.g. co-referent mention that attends to their antecedents). \cite{attention_clustering}, on the other hand, approaches the analysis of attention by means of an unsupervised clustering of attention heads. They find 4 groups of attention patterns and identify which of them have the greatest impact on accuracy. According to their research, association of a particular head to a group can change in the course of training and can also differ depending on the analysed dataset.

\section{Background}
\label{sec:background}

\subsection{BERT attention}
BERT architecture~\cite{BERT} consists of multiple attention layers, each of them containing multiple attention heads. An attention head takes as input a sequence of embeddings $t = [t_1, ..., t_n]$ corresponding to $n$ tokens of the input sentence. Those embeddings ($t_i$) are transformed into query, key, and value vectors ($q_i$, $k_i$, $v_i$) using $Q$, $K$ and $V$ matrices learned during the training process for each attention head separately. Each head computes attention weight $\alpha$ between all pairs of tokens according to Eq.~\ref{eq:attention_alpha}.

\begin{equation}
\label{eq:attention_alpha}
    \alpha_{ij} = \frac{exp(q_i^T k_j)}{\sum_{l=1}^n exp(q_i^T k_l)}
\end{equation}
The above stands for a softmax-normalized dot product between the query and key vectors.

\subsection{\textit{Mixup} augmentation} 
\textit{Mixup}~\cite{mixup} is a simple, yet effective, data-agnostic augmentation mechanism utilized in CV, that constructs synthetic training saples as a linear interpolation of the input images. Synthetic samples are constructed using Eqs.~\ref{eq:mixup_x} and~\ref{eq:mixup_y}.
\begin{equation}
\label{eq:mixup_x}
    \tilde{x} = \lambda x_i + (1-\lambda)x_j
\end{equation}
\begin{equation}
\label{eq:mixup_y}
    \tilde{y} = \lambda y_i + (1-\lambda)y_j
\end{equation}
where $x_i, x_j$ are two random samples from the training data, $y_i, y_j$ are their one-hot encoded labels, and $\lambda$ is the mixing ratio.

\section{Proposed approach}
\label{sec:approach}
\begin{figure*}[ht]
    \centering
    \includegraphics[width=\linewidth]{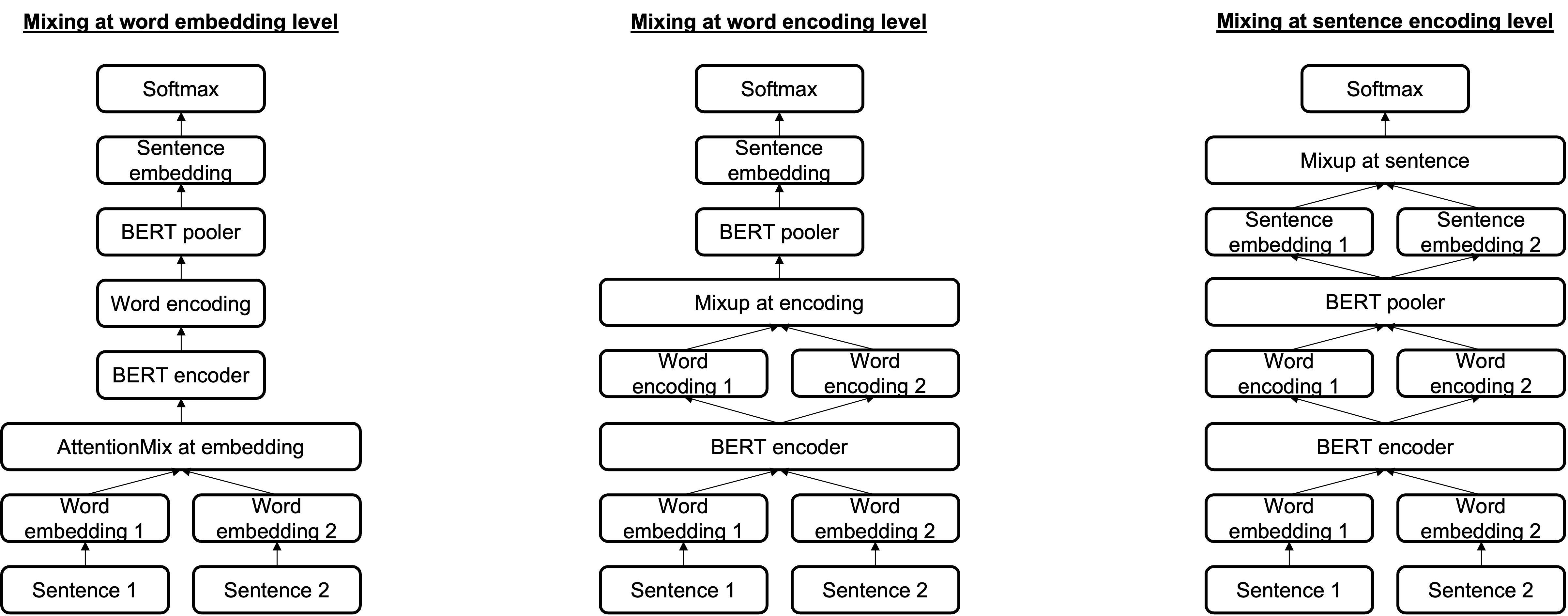}
    \caption{
    \textit{AttentionMix} at the word-embedding level proposed in this paper (left subfigure) and two \textit{Mixup} schemes: at the word-encoding level (middle) and at the sentence-encoding level (right). 
    }
    \label{fig:Mixing_options}
\end{figure*}

Mixing in BERT can be applied on various levels of the network:
\begin{itemize}
    \item At the word embedding level – Figure~\ref{fig:Mixing_options} (left) – the same word regardless of the context will have the same embedding vector.
    \item At the word encoding level – Figure~\ref{fig:Mixing_options} (middle) – the same word will have different embedding depending on the context.
    \item At the sentence embedding level – Figure~\ref{fig:Mixing_options} (right) – embedding takes place after vectors of individual words are aggregated to a sentence level.
\end{itemize}
The \textit{AttentionMix} method that we propose follows the first implementation concept.

\subsection{\textit{AttentionMix}}
\label{sec:AttentionMix}
\textit{AttentionMix} aims to utilize the information coming from attention heads (Eq.~\ref{eq:attention_alpha}) to guide the mixing process. Since attention information is relevant and applicable only on the word embedding and word encoding levels, the sentence embedding level will not be explored in this work (all token embeddings are already aggregated to sentence level embedding, at which stage utilization of attention assigned to individual tokens is impossible).

Furthermore, we focus on the augmentation at the word embedding level since the working hypothesis is that utilizing attention closer to the input and prior to the encoding stage (i.e. learning the context of each token and adjusting its embedding based on that) will lead to higher model's accuracy.

We explore various methods for utilizing attention information. Let's consider $L$ attention layers with $H$ heads each. Then, for each head $h \in H$ in layer $l \in L$ and each sentence $S$ the attention weight matrix $AW_{hl}(S)$ has the form:

\begin{equation}
\label{eq:attention_matrix}
    AW_{hl}(S) = [\alpha_{ij}]_{n\times n}
\end{equation}
where $n$ is the number of tokens in a sentence. $\alpha_{ij}$ represents the impact of token $\alpha_{ij}$ on the next-layer representation of the current token.

Based on $AW_{hl}(S)$, we calculate the relevance of each token in the sentence from the perspective of a single head (Eq.~\ref{eq:attention_head}) and the mean from all heads in a single layer (Eq.~\ref{eq:attention_layer}).
\begin{equation}
\label{eq:attention_head}
    B_{head_{hl}} = \frac{\sum_i \alpha_{ij}}{n}
\end{equation}
\begin{equation}
\label{eq:attention_layer}
    B_{layer_{l}} = \frac{\sum_{h=1}^H B_{head_{hl}}}{H}
\end{equation}
The above relevance of each token in a sentence is calculated for each observation (sentence) in the training dataset. For two pairs of (sentence, label): $(x_1, y_1)$ and $(x_2, y_2)$ the equations for creating a mixed sentence are as follows:
\begin{equation}
\label{eq:mixing_text_lambda_vector}
\lambda_{vector} = \frac{B^{1}}{B^{1}+B^{2}}
\end{equation}
\begin{equation}
\label{eq:mixing_text_x}
\tilde{x} = \lambda_{vector} \odot x_1 + (1-\lambda_{vector}) \odot x_2
\end{equation}
\begin{equation}
\label{eq:mixing_text_lambda}
\lambda_{label} = \frac{\sum \lambda_{vector}}{|\lambda_{vector}|}
\end{equation}
\begin{equation}
\label{eq:mixing_text_y}
\tilde{y} = \lambda_{label}\cdot y_1 + (1-\lambda_{label})\cdot y_2
\end{equation}
where $B^{1}$ and $B^{2}$ are the relevance vectors, calculated using either Eq.~\ref{eq:attention_head} or Eq.~\ref{eq:attention_layer}, for observations $(x_1, y_1)$ and $(x_2, y_2)$, respectively. $\lambda_{vector}$ is the mixing ratio vector used for token embedding mixing, $\lambda_{label}$ is the mixing ratio used to mix one-hot-encoded labels, and $|\lambda_{vector}|$ is the number of token relevance values. $\lambda_{vector}$ represents the importance of each individual token in a sentence and $\lambda_{label}$ is a single value (the mean of all $\lambda_{vector}$ elements) that is defines the relative degree to which each of the two one-hot-encoded vectors of labels contributes to the calculation of $\tilde{y}$ (Eq.~\ref{eq:mixing_text_y}).

Eqs.~\ref{eq:attention_head}-\ref{eq:mixing_text_y} formally describe the \textit{AttentionMix} algorithm that, to our knowledge, presents the first attempt to utilize attention coming from BERT to guide the \textit{Mixup} augmentation process.

\begin{algorithm*}[ht]
    \begin{algorithmic}[1]
    \For {$epoch=1,2,\ldots, max\_epoch$}
        \For {$batch=1,2,\ldots, max\_batch$}
            \State Extract token embeddings and attention weight matrices (\ref{eq:attention_matrix}) from BERT model
            \State Calculate relevance of each token according to either Eq.~\ref{eq:attention_head} or Eq.~\ref{eq:attention_layer}
            \State Create a shuffled copy of the above artefacts (embedding, relevance vector)
                \For {$observation=1,2,\ldots, max\_observation$}
                    \State Calculate the mixing ratio vector between $observation$ and its respective counterpart in a shuffled copy according to Eq.~\ref{eq:mixing_text_lambda_vector}
                    \State Calculate the contribution of each sentence from a pair according to Eq.~\ref{eq:mixing_text_lambda}
                    \State Create mixed sample based on Eq.~\ref{eq:mixing_text_x} and Eq.~\ref{eq:mixing_text_y}
                \EndFor
            \State Train model on $batch$ containing mixed samples
        \EndFor
    \EndFor
    \end{algorithmic}
\caption{\emph{AttentionMix}} 	
\label{alg:1}
\end{algorithm*}

\section{Experimental setup}
\subsection{Datasets}
We evaluate \textit{AttentionMix} on 3 sentence classification benchmark datasets, summarized in Table~\ref{tab:datasets_summary}.
\begin{itemize}
    \item SST – is the Stanford Sentiment Treebank dataset~\cite{sst} with fine-grained 5-level sentiment scale. 
    Note that in the literature, a simplified binary version of this data set is also considered. In the experiments, we chose the original non-binary setting with a 5-point sentiment scale.
    \item MR – is a Movie Review dataset with binary sentiment~\cite{mr}.
    \item IMDB -  Internet Movie Database is a much larger binary sentiment classification dataset for movie reviews \cite{imdb}.
\end{itemize}

\begin{table}
\caption{Train, validation, and test sizes of the benchmark datasets. 
}
\label{tab:datasets_summary}
\begin{center}
\begin{tabular}{|l|l|l|l|}
\hline
\textbf{Dataset} & \textbf{Train} & \textbf{Validation} & \textbf{Test} \\ \hline
SST              & 8544           & 1101                & 2210          \\ \hline
MR               & 8530           & 1066                & 1066          \\ \hline
IMDB             & 25000          & 10\% of Train       & 25000         \\ \hline
\end{tabular}
\end{center}
\end{table}


\subsection{Architecture and Experimental setup}
\label{arch_and_exp_setup}
All experiments were conducted using BERT network~\cite{BERT} with $12$ attention layers with $12$ heads each. The hidden size of the transformer block was equal to $768$. This version is commonly referred to as \textit{bert-base}.

For each method and each dataset, the experiment was conducted 3 times. Each run consisted in fine-tuning the BERT model initialized with \textit{bert-base-uncased} weights (the model pre-trained on English language with no distinction between small and capital letters) for a given dataset and method, for $100$ epochs with a learning rate of $1e^{-6}$ and dropout of $0.1$. All experiments were conducted on Nvidia A100 GPU 40GB. In the reminder of the paper, the word \textit{training} will refer to the above-described fine-tuning of the BERT model to a particular classification task.

In each experiment that utilizes \textit{Mixup}-based augmentation (i.e. all experiments except standard BERT training), the augmentation is applied during the training process to all training samples. Technically, \textit{AttentionMix} is implemented similarly to \textit{Mixup}, i.e. each batch of data is mixed with a shuffled version of the same batch, so the size of the training data does not change.

Each dataset was split into $3$ parts: train, val and test according to Table~\ref{tab:datasets_summary}. The results on the test part are presented, for the model that achieved the highest accuracy on the val part. The average outcome of 3 runs is reported.

\subsection{Benchmark methods and their hyperparameters}
We compare \emph{AttentionMix} with three baselines: (1) standard BERT training without \textit{Mixup}, referred to as vanilla approach, (2) adaptation of \textit{wordMixup}~\cite{mixup_for_text}, and (3) a special case of \textit{TMix}~\cite{tmix}, which we refer to as \textit{MixupEncoding}. The main difference compared to \textit{AttentionMix} is that both reference \textit{Mixup}-like augmentation methods do not use the guidance coming from the attention mechanism. Additionally, \cite{mixup_for_text} uses LSTM or CNN architecture instead of BERT, and the embeddings are utilized at the word level, not the token level. \textit{MixupEncoding} compared to \textit{TMix}~\cite{tmix} performs mixing after the BERT entire encoder, not at a randomly chosen hidden layer.

Apart from hyperparameters mentioned in Section \ref{arch_and_exp_setup}, the two \textit{Mixup}-based benchmark methods use one additional hyperparameter, i.e. a mixing ratio, sampled in the same way as in \textit{Mixup}~\cite{mixup}.

\section{Experimental results and analysis}

\subsection{Stanford Sentiment Treebank dataset} 

The results for the SST dataset are presented in Table~\ref{tab:top_SST}. The vanilla BERT method reached $51.17\%$ and was inferior to both other benchmarks that utilize \textit{Mixup} in the training process (\textit{wordMixup} and \textit{MixupEncoding}).

In all \textit{AttentionMix} experiments presented in Table~\ref{tab:top_SST}, higher mean accuracy than the vanilla approach was achieved, and in all but one of them the \textit{AttentionMix} results exceeded all $3$ benchmark approaches.

\begin{figure*}[htb!]
    \centering
    \includegraphics[width=0.31\linewidth]{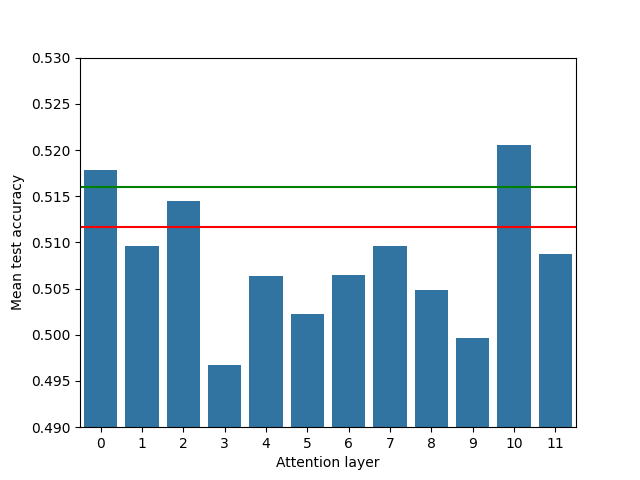}
    \ \ 
    \includegraphics[width=.31\linewidth]{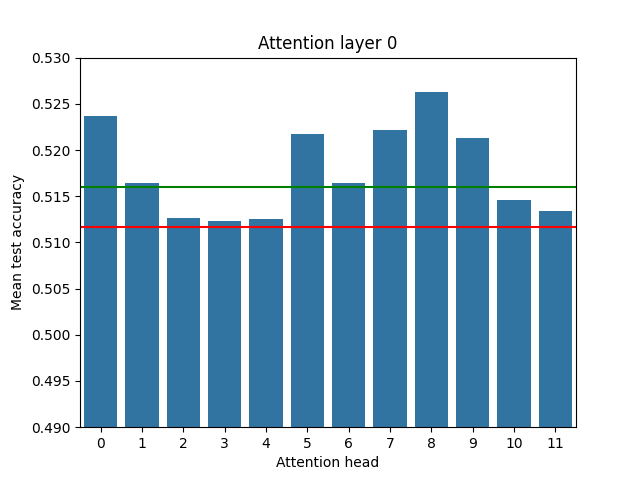}
    \ \ 
    \includegraphics[width=.31\linewidth]{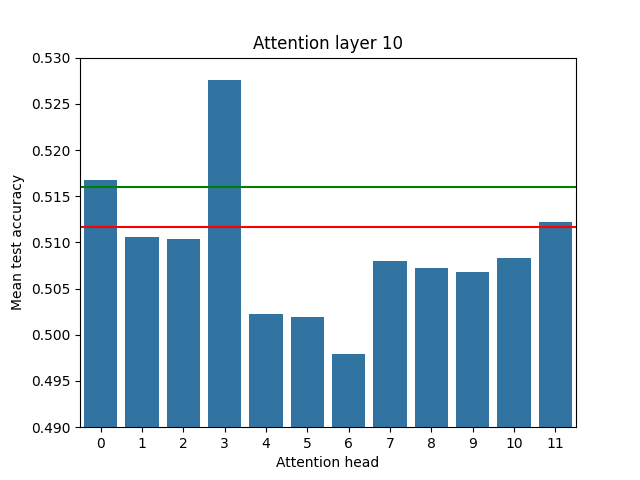}
    \caption{SST dataset. The average test accuracy for experiments utilizing mean attention from: all heads in a layer (Eq.~\ref{eq:attention_layer}) -- \textit{left subfigure}, single heads (Eq.~\ref{eq:attention_head}) in layer 0 -- \textit{middle subfigure}, single heads (Eq.~\ref{eq:attention_head}) in layer 10 -- \textit{right subfigure}. The red line indicates the vanilla BERT accuracy, and the green one, the best accuracy achieved by the \textit{Mixup} benchmark methods. All experiments were repeated 3 times.}
    \label{fig:SST-main}
  \end{figure*}

More detailed results are depicted in Figure~\ref{fig:SST-main}. The left subfigure presents the average accuracy when all heads in a given layer are utilized, and the middle and right subfigures are deep dives into the average accuracy when single heads in the top performing layers ($0$ and $10$, respectively) are considered. 

The results presented in Figure~\ref{fig:SST-main} show that the use of other than top-$3$ attention layers in the augmentation process clearly deteriorates the accuracy on the test set. Furthermore, when looking at the middle and right subfigures, the highest result among individual heads is achieved by a head from layer $10$, the learning in layer $0$ is more uniform and there are more ``strong'' heads in this layer. In layer $0$ the use of any individual head results in higher accuracy than the standard BERT training and for $6$ out of $12$ heads it excels all competitive methods.

On the contrary, a closer look at layer $10$ (Figure~\ref{fig:SST-main} (right)) shows that there are only $2$ heads ($0$ and $3$) with the results higher than all benchmarks and just $3$ ($0$, $3$ and $11$) with the accuracy higher than the standard BERT training.

\begin{table}[!htb] \footnotesize
\caption{SST dataset. Comparison of the average results of 3 benchmark methods and \emph{AttentionMix}. For \emph{AttentionMix}, two attention levels are considered: the \textbf{layer level} – (all heads from a layer) and the \textbf{head level} – (a single head within all layers). In each case, the results of the $3$ best performing configurations are presented.
The details are depicted in Figure~\ref{fig:SST-main}.
All experiments were repeated $3$ times.
}
\label{tab:top_SST}
\begin{center}
\begin{tabular}{|l|l|c|c|c|c|}
\hline
                    Approach &   \multicolumn{2}{c|}{Attention}  &  \multicolumn{2}{c|}{Accuracy} \\ 
                     &   layer &   head &   mean &   std \\
\hline
                       standard training &              --- &             --- &          51.17 &          0.97 \\
            \textit{wordMixup} &              --- &             --- &          51.60 &          0.18 \\
              \textit{MixupEncoding} &              --- &             --- &          51.30 &          1.13 \\
      \emph{AttentionMix} &               10 &             all &          52.05 &          0.86 \\
     \emph{AttentionMix} &                0 &             all &          51.78 &          0.23 \\
     \emph{AttentionMix} &                2 &             all &          51.45 &          0.21 \\
\emph{AttentionMix} &               10 &               3 &          52.76 &          0.58 \\
 \emph{AttentionMix} &                0 &               8 &          52.62 &          0.21 \\
 \emph{AttentionMix} &                0 &               0 &          52.37 &          0.26 \\
\hline
\end{tabular}
\end{center}
\end{table}

\subsection{Movie Review dataset}
Table~\ref{tab:top_MR} presents the results for the MR dataset. The vanilla benchmark reached $85.14\%$. Clearly, there is a substantial difference in accuracy between MR and SST, due to the fact that  MR is a binary — hence potentially much less complex – dataset. Interestingly, both methods that utilize \textit{Mixup} in the training process (\textit{wordMixup} and \textit{MixupEncoding}) achieved lower accuracy, which is in line with the results reported in \cite{mixup_for_text_2}. 

\begin{figure*}[htb!]
    \centering
    \includegraphics[width=.31\linewidth]{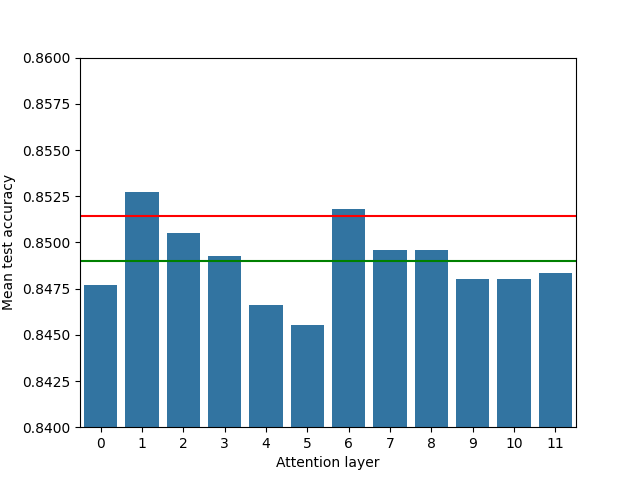}
    \ \ 
    \includegraphics[width=.31\linewidth]{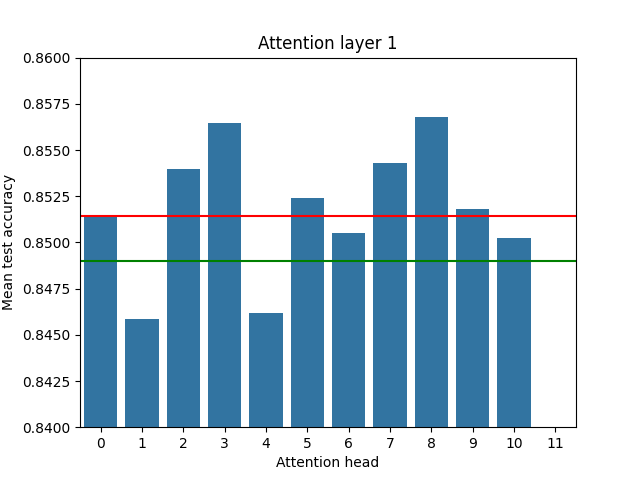}
    \ \ 
    \includegraphics[width=.31\linewidth]{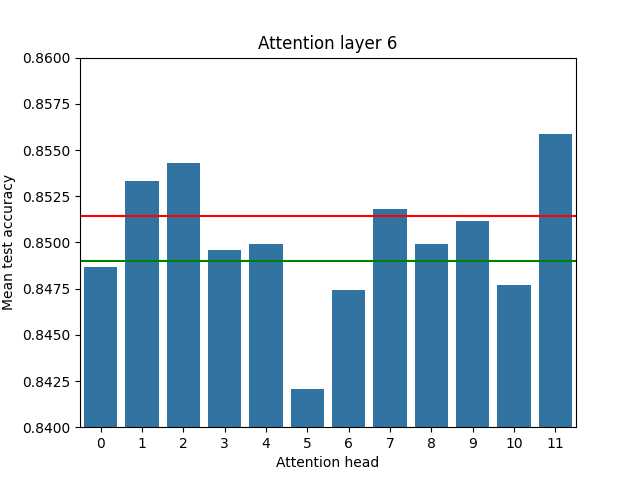}
    \caption{MR dataset. The average test accuracy for experiments utilizing mean attention from: all heads in a layer (Eq.~\ref{eq:attention_layer}) -- \textit{left subfigure}, single heads (Eq.~\ref{eq:attention_head}) in layer 1 -- \textit{middle subfigure}, single heads (Eq.~\ref{eq:attention_head}) in layer 6 -- \textit{right subfigure}. The red line indicates the 
    vanilla BERT accuracy,
    and the green one, the best accuracy achieved by the \textit{Mixup} benchmark methods. All experiments were repeated 3 times.}
    \label{fig:MR-main}
  \end{figure*}

In $5$ out of $6$ experiments, \textit{AttentionMix} achieved higher mean accuracy than the vanilla training. The experiments utilizing attention from particular layers achieved the highest results. 
Similarly to SST, detailed results are available in Figure~\ref{fig:MR-main}. The left subfigure presents the average accuracy resulting from utilizing all heads in a given layer and middle and right subfigures are head related outcomes, i.e. the average accuracy when single heads in the top performing layers (layer 1 and layer 6, respectively) are used. 

In the left subfigure, only $2$ attention layers achieved the test accuracy higher than the best benchmark approach, which in this case is standard BERT training. When looking at middle and right subfigures a trend similar to SST dataset can be observed, i.e. generally more heads that outperform the best benchmark approach can be found in earlier layers than in the later ones. In the middle subfigure, which refers to layer $1$, there are $7$ heads with test accuracy exceeding all benchmark methods, while for layer $6$ (right subfigure) there are only $4$ such heads.

\begin{table}[!htb]\footnotesize
\caption{MR dataset. Comparison of the average results of 3 benchmark methods and \emph{AttentionMix}. For \emph{AttentionMix}, two attention levels are considered: the \textbf{layer level} – (all heads from a layer) and the \textbf{head level} – (a single head within all layers). In each case, the results of the 3 best performing configurations are presented. The details are depicted in Figure~\ref{fig:MR-main}.
All experiments were repeated 3 times.
}
\label{tab:top_MR}
\begin{center}
\begin{tabular}{|l|l|l|l|l|}
\hline
                      Approach &   \multicolumn{2}{c|}{Attention}  &  \multicolumn{2}{c|}{Accuracy}  \\
                  &   layer &   head &   mean &   std \\
\hline
                         standard training &              --- &             --- &          85.14 &          0.61 \\
             \textit{wordMixup} &              --- &             --- &          84.90 &          0.86 \\
               \textit{MixupEncoding} &              --- &             --- &          85.08 &          0.41 \\
     \emph{AttentionMix} &                1 &             all &          85.27 &          0.47 \\
     \emph{AttentionMix} &                6 &             all &          85.18 &          0.43 \\
     \emph{AttentionMix} &                2 &             all &          85.05 &          0.24 \\
\emph{AttentionMix} &                1 &               8 &          85.68 &          0.53 \\
 \emph{AttentionMix} &                1 &               3 &          85.65 &          0.28 \\
\emph{AttentionMix} &                6 &              11 &          85.58 &          0.11 \\
\hline
\end{tabular}
\end{center}
\end{table}

  \begin{figure*}[htb!]
    \centering
    \includegraphics[width=.31\linewidth]{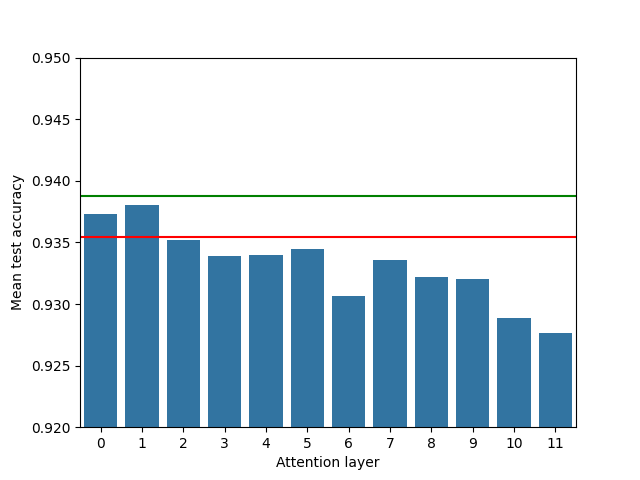}
    \ \ 
    \includegraphics[width=.31\linewidth]{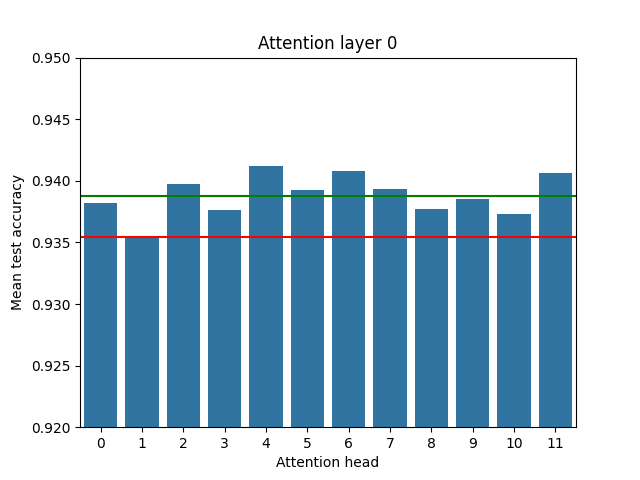}
    \ \ 
    \includegraphics[width=.31\linewidth]{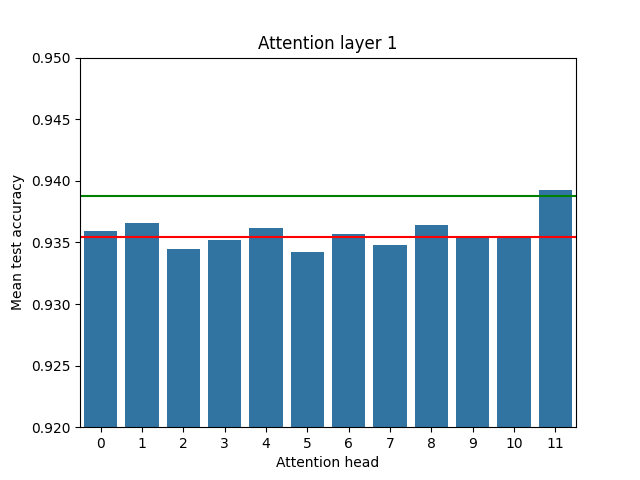}
    \caption{IMDB dataset. The average test accuracy for experiments utilizing mean attention from: all heads in a layer (Eq.~\ref{eq:attention_layer}) -- \textit{left subfigure}, single heads (Eq.~\ref{eq:attention_head}) in layer 0 -- \textit{middle subfigure}, single heads (Eq.~\ref{eq:attention_head}) in layer 1 -- \textit{right subfigure}. The red line indicates the 
    vanilla BERT accuracy,
    and the green one, the best accuracy achieved by the \textit{Mixup} benchmark methods. All experiments were repeated 3 times.}
    \label{fig:IMDB-main}
  \end{figure*}

\subsection{Internet Movie Database dataset}
The results for the IMDB dataset are shown in Table~\ref{tab:top_IMDB}. The vanilla BERT approach achieved $93.54\%$ and was weaker than both \textit{Mixup}-related benchmarks. IMDB and MR datasets, both refer to binary sentiment classification, though IMDB has 3 times more training samples, which is the most probable reason of higher accuracy on this dataset in all experiments, compared to MR.

\textit{AttentionMix} achieved higher mean accuracy than the vanilla training in 5 out of 6 experiments. Additionally, all \textit{AttentionMix} experiments utilizing individual heads (the last $3$ rows of Table~\ref{tab:top_IMDB}) produced  the results exceeding all $3$ benchmark approaches. Head-related results are illustrated in Figure~\ref{fig:IMDB-main}. The left subfigure presents the average accuracy when all heads in a given layer are used, and middle and right subfigures show the average accuracy of single heads in the best performing layers (0 and 1, respectively). 

It can be seen in Figure~\ref{fig:IMDB-main} that the use of other than top-$2$ attention layers in the augmentation process deteriorates the accuracy on the test set below the accuracy of the benchmark methods. The highest result among individual heads is achieved by the head from layer $0$ (middle subfigure). Similarly to SST, learning in this layer is more uniform, with high number of effective heads. The use of any individual head from layer $0$ results in higher accuracy than BERT and half of the heads excel all $3$ competitive approaches.

On the contrary, in Figure~\ref{fig:IMDB-main} (right), that refers to layer $1$, only one head (number $11$) is superior to all benchmark approaches and half of the heads exceed the BERT benchmark.

In summary, there are two main conclusions that can be drawn from the presented experiments on the three sentence classification datasets. Firstly, if a decision about \emph{AttentionMix} settings needs to be made without the possibility of checking different hyperparameter configurations (e.g. due to limited computational resources), it is advised to use the vector of relevance that is extracted from a single head since such vectors yielded higher performance in our experiments. Secondly, this single head should rather be selected among heads belonging to the initial layers. It stems from the experiments, that in these earlier layers there are more heads that outperform benchmark results, albeit with a word of warning that with such a head selection strategy the overall best performing head can potentially be omitted (e.g. head $3$ from layer $10$ in SST experiments).

\begin{table}[!htb]\footnotesize
\caption{IMDB dataset. Comparison of the mean results of 3 benchmark methods and \emph{AttentionMix}. For \emph{AttentionMix}, two attention levels are considered: the \textbf{layer level} – (all heads from a layer) and the \textbf{head level} – (a single head amongst all layers). In each case, the results of the 3 best performing configurations are presented.
The details are depicted in Figure~\ref{fig:IMDB-main}.
All experiments were repeated 3 times. 
}
\label{tab:top_IMDB}
\begin{center}
\begin{tabular}{|l|l|l|l|l|}
\hline
               Approach &   \multicolumn{2}{c|}{Attention}  &  \multicolumn{2}{c|}{Accuracy}  \\
                    &   layer &   head &   mean &   std \\
\hline
                       standard training &              --- &             --- &          93.54 &          0.21 \\
            \emph{wordMixup} &              --- &             --- &          93.88 &          0.14 \\
               \emph{MixupEncoding} &              --- &             --- &          93.72 &          0.16 \\
     \emph{AttentionMix} &                1 &             all &          93.80 &          0.06 \\
     \emph{AttentionMix} &                0 &             all &          93.73 &          0.29 \\
     \emph{AttentionMix} &                2 &             all &          93.52 &          0.09 \\
 \emph{AttentionMix} &                0 &               4 &          94.12 &          0.03 \\
 \emph{AttentionMix} &                0 &               6 &          94.08 &          0.18 \\
 \emph{AttentionMix} &                0 &              11 &          94.06 &          0.11 \\
\hline
\end{tabular}
\end{center}
\end{table}

\subsection{Computational complexity}
\textit{AttentionMix} training requires 2 additional operations compared to \textit{wordMixup}: partial forward pass (the sentence is passed through encoder to get attentions) and calculation of $\lambda$ vector (non-computationally intensive mean). The training time overhead is 
around $26\%$.
The inference time is 
exactly the same for both approaches.

\subsection{Ablation study}
We further investigated why certain information coming from attention weight matrices results in higher accuracy boost. Since the three datasets consist of various types of sentiment analysis, we hypothesized that certain parts of speech may have higher impact on sentiment classification than others. Specifically, our assumption was that adjectives, adverbs, and verbs could possibly be more indicative for the sentiment class prediction, since the sentiment is usually reflected by the statements like:
\begin{itemize}
    \item love, like, hate – verbs
    \item fantastic, disappointing – adjectives
    \item quite, very, extremely – adverbs
\end{itemize}

\begin{figure}[htb!]
    \centering
    \includegraphics[width=1\linewidth]{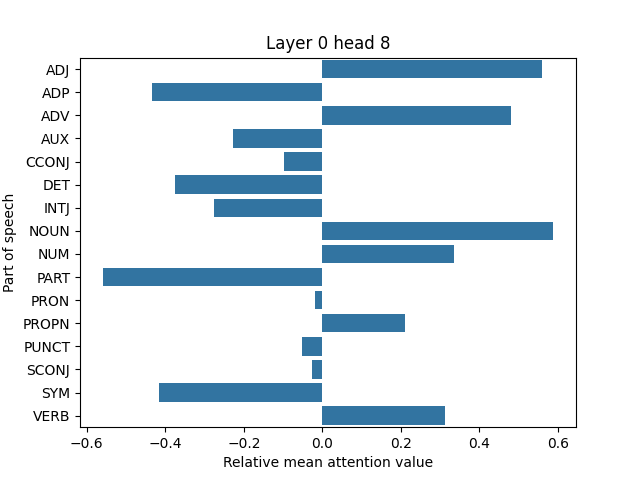}
    \includegraphics[width=1\linewidth]{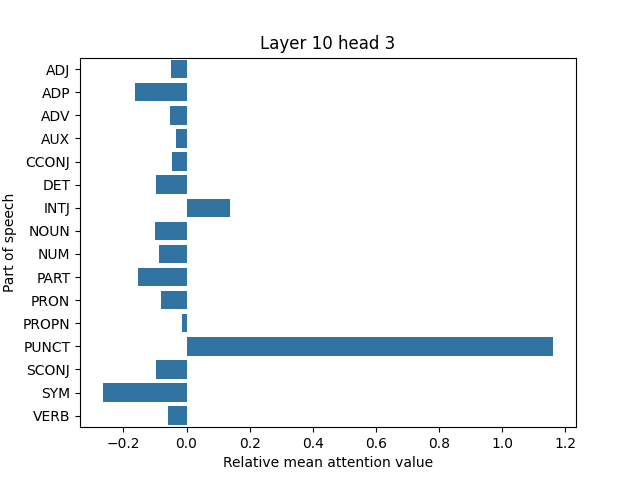}
        \caption{SST dataset. The mean attention value assigned to each part of speech for a given attention head, relative to the mean attention value of the head after training on SST. 
        The abbreviations stand for: ADJ – adjective, ADP – adposition, ADV – adverb, AUX – auxiliary, CONJ - conjunction, CCONJ - coordinating conjunction, DET – determiner, INTJ – interjection, NOUN – noun, NUM - numeral, PART – particle, PRON - pronoun, PROPN - proper noun, PUNCT - punctuation, SCONJ - subordinating conjunction, SYM - symbola and VERB - verb.}
\label{fig:SST_mean_att_0_8_epoch_last}
\end{figure}

For the SST dataset, this hypothesis was confirmed only for some relevance vectors derived from attention information. Figure~\ref{fig:SST_mean_att_0_8_epoch_last} shows the mean attention given to a certain part of speech by attention head 8 in layer 0 and attention head 3 in layer 10, whose usage resulted in the two best performing models. For head 8 in layer 0 indeed high attention is given to adjectives, adverbs, and verbs, but for attention head 3 in layer 10 very high attention is given to punctuation. This phenomenon of high punctuation-related attention has been previously observed in~\cite{annotation_analysis}.

\begin{figure}[ht!]
    \centering
    \includegraphics[width=1\linewidth]{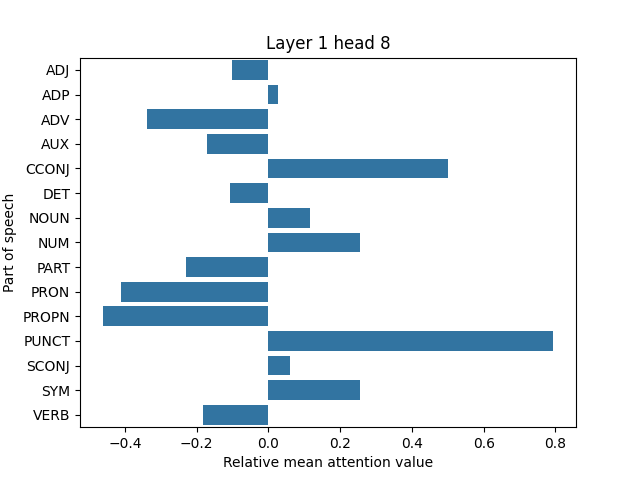}
    \includegraphics[width=1\linewidth]{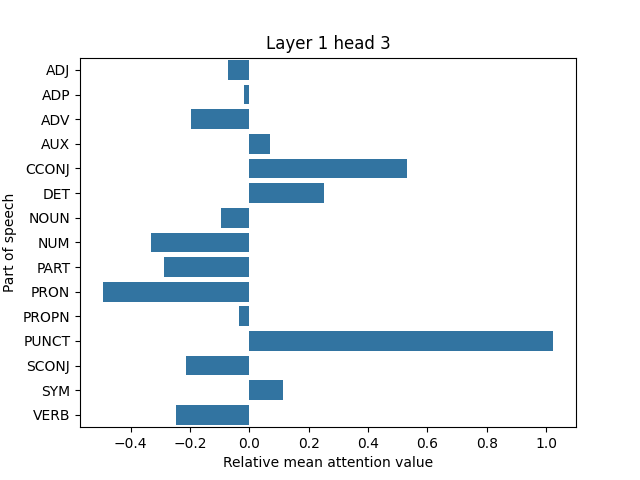}
    \caption{MR dataset. The mean attention value assigned to each part of speech for a given attention head, relative to the mean attention value of the head after training on MR.
    For the list of abbreviations, please refer to the caption of Figure~\ref{fig:SST_mean_att_0_8_epoch_last}.}
    \label{fig:MR_mean_att_1_8_epoch_last}
\end{figure}

For the MR dataset, the relative mean attention assigned to each part of speech for the two top-performing heads is presented in Figure~\ref{fig:MR_mean_att_1_8_epoch_last}. Surprisingly, for both heads the attention points to seemingly unintuitive direction, at least for humans. The highest attention is attributed to PUNCT and CCONJ which for this dataset cover mostly citation symbols, various types of brackets, a comma, a dot, a question mark, an exclamation mark, and 3 common words: \textit{and}, \textit{but} and \textit{or}.

\begin{figure}[ht!]
    \centering
    \includegraphics[width=1\linewidth]{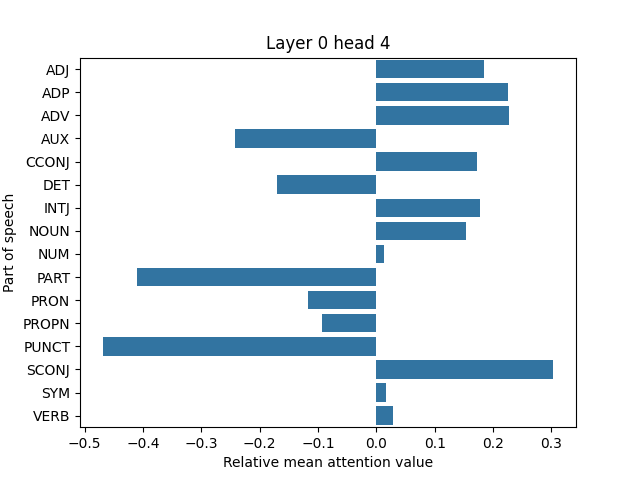}
    \includegraphics[width=1\linewidth]{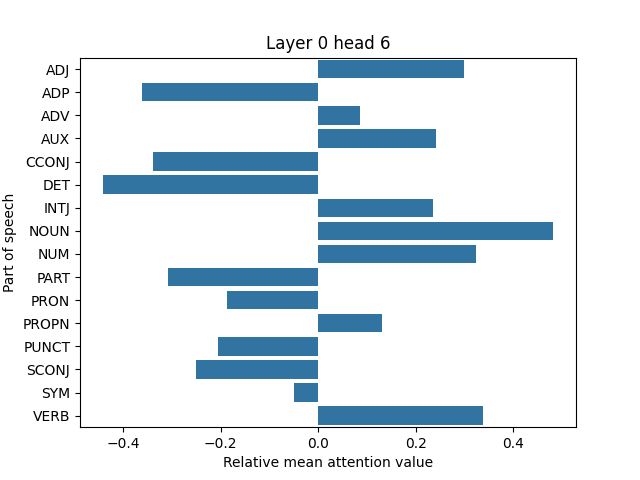}
    \caption{IMDB dataset. The mean attention value assigned to each part of speech for a given attention head, relative to the mean attention value of the head after training on IMDB.
    For the list of abbreviations, please refer to the caption of Figure~\ref{fig:SST_mean_att_0_8_epoch_last}.}
    \label{fig:IMDB_mean_att_0_4_epoch_last}
\end{figure}

For the IMDB dataset, the two best performing heads belong to layer $0$ (heads $4$ and $6$). The relative mean attention assigned to each part of speech for these two heads is depicted in Figure~\ref{fig:IMDB_mean_att_0_4_epoch_last}. Similarly to the SST results for layer 0 head 8, the original hypothesis stands for both heads. Among other parts of speech, head $4$ attends to adjectives and adverbs, and head $6$ to adjectives and verbs.

In summary, across all 3 datasets, in $3$ out of $6$ top performing heads the attention is given to adjectives, adverbs, and verbs. In the remaining $3$ cases, the attention choices are less intuitive. 

In essence, based on the results presented in Tables~\ref{tab:top_SST},~\ref{tab:top_MR} and~\ref{tab:top_IMDB} we conclude that using information from the attention mechanism is helpful in creating augmented samples. At the same time, a detailed explanation of the reasons for particular attention focus in certain cases requires further studies.

The existence of unintuitive attention heads has been already observed in the literature~\cite{attention_clustering, attention_head_hypothesis}. The first paper, besides other topics, studies the so-called vertical attention heads that attend mostly to dots, comas and BERT special tokens. The other work refers to the above as delimiter heads, and quantifies their high prevalence at the level of $73.43\%$, stating that they naturally coexist with other functions of the head.

The ablation experiments presented in the paper concentrated on sentence $x_1$, based on which the mixing ratio vector ($\lambda_{vector}$ in Eq.~\ref{eq:mixing_text_lambda_vector}) is calculated. Another possible path, though harder to implement, is analysis of the attention focus based on sentence $x_2$. We plan to investigate this path in our future work.

\section{Concluding remarks}
Inspired by the success of \textit{Mixup} augmentation in the CV domain, we introduce a \textit{Mixup}-related augmentation method in the context of text classification. Unlike previous \textit{Mixup}-based approaches devoted to the NLP domain, we propose to use a guided mixing approach, and towards this end utilize BERT attention information as the source of guidance for the augmentation process. We show  empirically that the proposed method, \emph{AttentionMix}, outperforms the vanilla BERT approach and two \textit{Mixup}-based benchmark methods used for comparison. The results support the effectiveness of the use of guided attention-based mixing in the NLP domain.

In future work, we plan to search for an automated method of selecting the most relevant attention information for a given dataset. This goal is associated with the well-known difficulty in interpreting the information coming from the attention heads~\cite{annotation_analysis,attention_16_heads_better_than_1}. A phenomenon that is far from fully understood. 

Another research direction is evaluation of the \textit{AttentionMix} efficacy when applied on the word encoding level – as opposed to the current word embedding level implementation.

\bibliographystyle{splncs04}
\bibliography{manuscript}

\end{document}